%% file: main.tex
 \documentclass[sigconf]{acmart}

\usepackage{comment}

\usepackage{booktabs}
\usepackage[utf8]{inputenc}
\usepackage[ruled,vlined]{algorithm2e}
\usepackage{microtype}
\usepackage{graphicx}
\usepackage{paralist}
\usepackage{tabularx}
\usepackage{balance}
\usepackage{multicol}
\usepackage{multirow}
\usepackage{pbox}
\usepackage{enumitem}
\usepackage{colortbl}
\usepackage{pifont}
\usepackage{xspace}
\usepackage{url}
\usepackage{tikz}
\usepackage{rotating}
\usepackage{balance}
\usepackage{float}
\usepackage[TABBOTCAP]{subfigure}
\usepackage{tcolorbox}
\usepackage{stackengine}
\usepackage{paralist}
\usepackage{xcolor}
\usepackage{listings}

\lstdefinestyle{jsonstyle}{
    basicstyle=\ttfamily\tiny,
    breaklines=true,
    breakatwhitespace=false,
    columns=fullflexible,
    keepspaces=true,
    showstringspaces=false,
    frame=single,
    numbers=none
}

\makeatletter
\newcommand{\linebreakand}{%
  \end{@IEEEauthorhalign}
  \hfill\mbox{}\par
  \mbox{}\hfill\begin{@IEEEauthorhalign}
}
\makeatother

\usepackage{listings}

\input{macro}

\begin{document}


\title{AIGen: Automating AI Bill of Materials Generation Through Hybrid MLOps Integration}


\author{Federica Pepe}
\affiliation{
  \institution{University of Sannio}
  \city{Benevento}
  \country{Italy}
}
  \email{f.pepe8@studenti.unisannio.it}

\author{Daniele Bifolco}
\affiliation{
  \institution{University of Sannio}
  \city{Benevento}
  \country{Italy}
}
\email{d.bifolco@studenti.unisannio.it}

\author{Costantino Martignetti}
\affiliation{
  \institution{University of Sannio}
  \city{Benevento}
  \country{Italy}
}
  \email{costa.martignetti@gmail.com}

\author{Aureliano D'Amici}
\affiliation{
  \institution{Smart Shaped s.r.l.}
  \city{Pescara}
  \country{Italy}
}
  \email{aureliano.damici@smartshaped.com}

\author{Fabiano Izzo}
\affiliation{
  \institution{Smart Shaped s.r.l.}
  \city{Pescara}
  \country{Italy}
}
  \email{fabiano.izzo@smartshaped.com}

\author{Damian A. Tamburri}
\affiliation{
  \institution{University of Sannio}
  \city{Benevento}
  \country{Italy}
}
  \email{datamburri@unisannio.it}

\author{Massimiliano Di Penta}
\affiliation{
  \institution{University of Sannio}
  \city{Benevento}
  \country{Italy}
}
  \email{dipenta@unisannio.it}


\begin{abstract}
The responsible development and deployment of artificial intelligence (AI) systems requires rigorous documentation of their constituent artifacts, e.g., datasets, model weights, training pipelines, and runtime dependencies. Although the Software Package Data Exchange (SPDX) 3.0 standard introduced native support for AI and dataset profiles, practical tooling capable of generating standards-compliant AI Bills of Materials (AIBoMs) in an automated and extensible manner remains scarce. This paper presents AIGen, a modular AIBoM generator that produces machine-readable, interoperable inventories of AI system components that comply with the SPDX 3.0 AI profile.
AIGen works on top of the MLflow MLOps framework and combines mining heuristics with Large Language Models to generate AIBoMs. A plugin interface allows practitioners to extend the tool with domain-specific collectors without modifying the core codebase, supporting heterogeneous AI frameworks such as Hugging Face, PyTorch, and TensorFlow. 
AIGen is designed to facilitate compliance with the European Union AI Act, the NIST AI Risk Management Framework, and ISO/IEC 42001, providing a concrete, reusable foundation for transparent, accountable AI supply chain governance. \\
\underline{Tool URL: \url{https://github.com/danielebifolco/AIGen}}\\
\underline{Tool Video: \url{https://youtu.be/\_nAbXDWfVL4}}

\end{abstract}



\maketitle
\renewcommand{\shortauthors}{F.Pepe, D.Bifolco, C.Martignetti, A.D'Amici, F.Izzo, D.A.Tamburri, M.Di Penta}


\section{Introduction}
\label{sec:intro}
\input{intro}

\section{Architecture}
\label{sec:architecture}
\input{architecture}

\section{Implementation and Usage Workflow}
\label{sec:implementation}
\input{implementation}

\section{Preliminary Evaluation}
\label{sec:eval}
\input{eval}

\section{Related Work}
\label{sec:related}
\input{related}

\section{Conclusions}
\label{sec:conc}
\input{conclusion}

\section*{Data Availability Statement}
The study replication package, containing raw data, scripts, and the tool, is available online~\cite{replicationPKGZenodo}.

\balance

\bibliographystyle{ACM-Reference-Format}

\bibliography{bib.bib}

\end{document}

%% file: intro.tex
The concept of a Bill of Materials (BoM) has so far served as a foundational tool in manufacturing and software development \cite{xia2023empirical,stalnaker2024boms}, providing a structured inventory of all components required to build a product. In the context of AI, this concept has been extended to become \textit{AI Bill of Materials} (AIBoM): a comprehensive, machine-readable record that catalogs every artifact involved in the development, training, and deployment of an AI system. Unlike traditional Software BoMs (SBoMs), which track libraries and source code dependencies, AIBoMs must account for the distinctive nature of AI systems, where behavior is determined as much by training data as by code itself. An AIBoM captures model metadata (architecture, versioning, and provenance), dataset information (sources, licensing, and preprocessing steps), training pipeline configurations, and runtime environment specifications. 
This comprehensive collection of information creates a complete and auditable fingerprint of the AI system throughout its lifecycle~\cite{stewartSeip26, bennet2025spdx}. In some domains, these features are a must-have, leading to severe capital losses or legal disputes if non-compliance occurs. 

For example, suppose H GmbH is a fictional German health-tech firm with €420M revenue; H deploys a lung CT scanner machine accompanied by specialized diagnostic AI---costing 500k a piece---across 47 hospitals. No AIBoM exists. A performance regression triggers a BfArM\footnote{\url{www.bfarm.de}} audit, unraveling an unlicensed training dataset and a complete absence of documentation. Quantitative impacts according to EU regulations would totalize around 26.2M EUR direct + 58M EUR market cap loss\footnote{\url{artificialintelligenceact.eu/article/99}}.

On the one hand, standardization efforts for AIBoMs have advanced significantly in recent years, with two major initiatives shaping the field. The first is SPDX~3.0, maintained by the Linux Foundation, which extended the established software supply chain standard to natively represent AI-specific elements such as datasets and iterative training artifacts~\cite{bennet2025spdx, 11025719, stewartSeip26}. The second is the OWASP CycloneDX ML-BoM specification, an Ecma International standard that enables organizations to document model cards, training parameters, ethical considerations, and data provenance within a unified, interoperable format~\cite{cyclonedx2023mlbom}. Both frameworks treat the AIBoM as a \textit{living document}---one initiated during system planning, continuously updated through training and validation, and carried into production alongside the model. Emerging tooling such as AIGen has further demonstrated that AIBoM generation can be automated with negligible performance overhead, using cryptographically signed attestations to guarantee the integrity of all captured artifacts~\cite{vandendriessche2026aibomgen}.

On the other hand, the growing adoption of AIBoM systems is driven not only by technical necessity but also by a rapidly evolving regulatory landscape; scalable and reliable AIBoM composition, maintenance, and evolution remain elusive.

This paper describes AIGen, a tool that streamlines AI documentation generation within standard MLOps frameworks through a hybrid extraction approach. 
AIGen is built on top of the MLflow \cite{mlflow} MLOps platform and is therefore designed to be scalable and reliable, leveraging it to directly extract structured data, such as hyperparameters and metrics, while using a plugin-based architecture to extend data collection to other sources. For unstructured data, such as use cases and logs, AIGen combines data mining heuristics and Large Language Models (LLMs) to assess source code and documentation. This integrated strategy automates data collection, significantly reducing the likelihood of errors from manual data entry. To address the variability typically found in industrial software, our framework employs a modular architecture via \textit{Builders}---constructor structures that separate data extraction from output generation--- ensuring the core system does not depend on specific MLOps tools or data sources. Although AIGen presently supports SPDX 3.0, the architecture is designed to accommodate new formats or standards with minimal effort.

%% file: architecture.tex
\begin{figure} 
    \centering
    \small
    \includegraphics[width=\columnwidth]{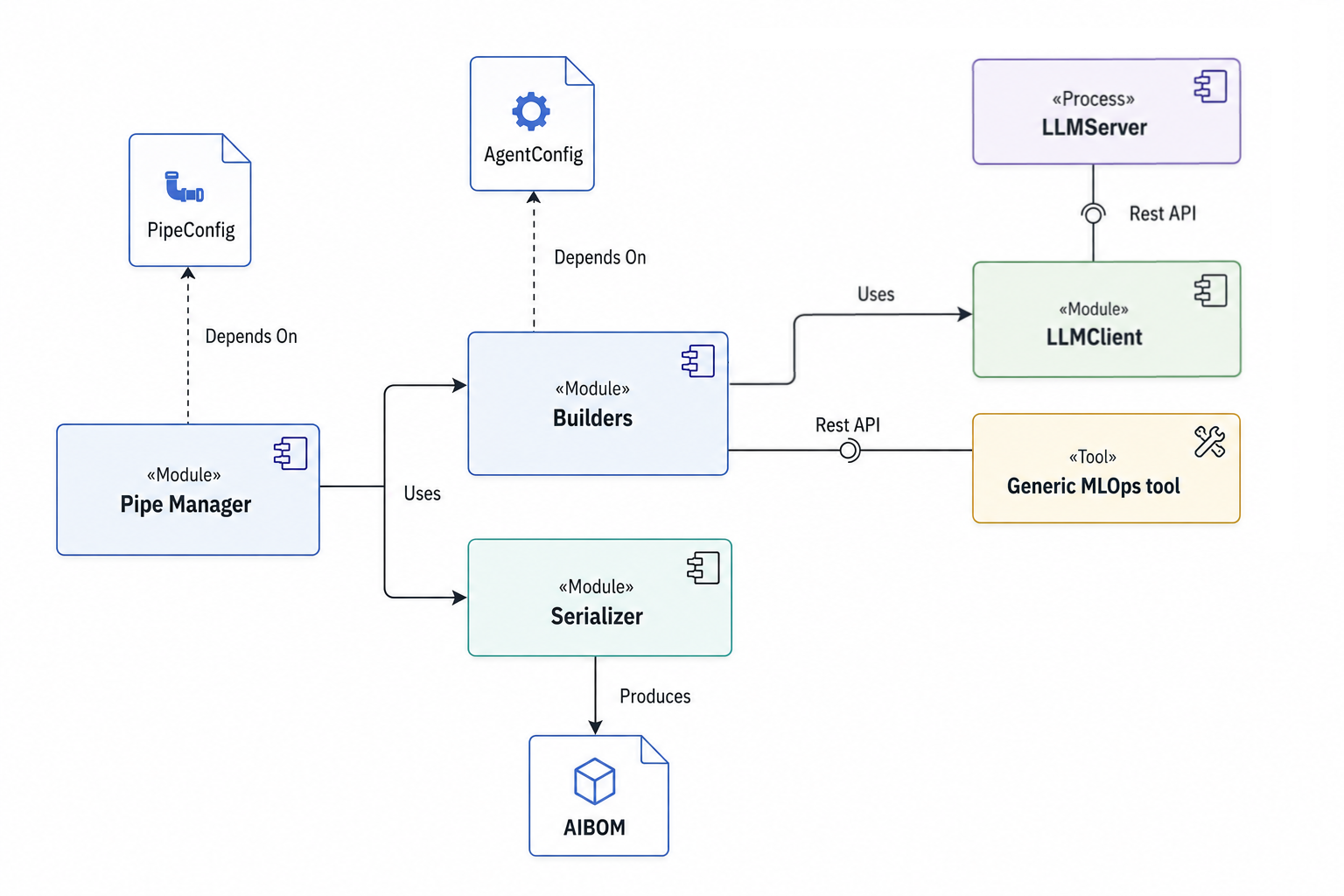}
    \caption{Architecture and operational pipeline of AIGen}
    \label{fig:architecture} 
\end{figure}

AIGen operates using a five-phase pipeline. It begins with \textit{Configuration} and \textit{Resource Initialization}, 
where the system dynamically identifies and loads the required modules (Builders, Teams, Goals) tailored to the specific AI project.
In the \textit{Data Retrieval} phase, active Builders gather metadata via APIs or LLMs, which are then serialized into the designated format. Lastly, a \textit{Validation} phase confirms that the output complies with formal AIBoM specifications.

Figure~\ref{fig:architecture} shows the AIGen's architecture and the process flow from setup to verified output.

\textit{\textbf{PipeManager}}

Is the Java-based engine that oversees pipeline execution. It handles the runtime lifecycle, sets up document metadata (such as authorship), and organizes data retrieval tasks using a priority-driven approach. To promote extensibility, it employs Java reflection to identify and instantiate \textit{Builders} at runtime. This feature allows developers to introduce new MLOps tools or data sources by implementing the class without modifying the core codebase, ensuring adaptability to evolving infrastructure without recompilation.

\textit{\textbf{Builders, Teams, and Goals}}\\
Three complementary entities manage data extraction:
\textbf{Builders:} 
Specialized units that extract metadata from diverse sources (\eg MLflow or GitHub). Their design allows developers to implement only the logic required for specific data types, ensuring seamless extensibility for new integrations.
Our architecture currently includes four reference implementations:
(i) \textit{MLflow Builder:} Utilizes REST APIs to gather structured metadata (including versions, hyperparameters, metrics) as well as custom SPDX tags.
(ii) \textit{GitHub Builder:} Connects with the \textit{LLMClient} to analyze training scripts and documentation, producing high-level summaries through LLM evaluation.
(iii) \textit{Kaggle Builder:} Concentrates on data provenance; it reviews dataset headers and metadata to categorize data types (such as tabular and image) without handling the raw files.
(iv) \textit{Hugging Face Hub Builder:} Retrieves model repository metadata, model-card content, license tags, framework tags, revision information, and download locations from Hugging Face Hub.
\textbf{Teams:} Logical groupings of Builders that consolidate data for a single SPDX component. This allows combining disparate sources, such as VCS licensing and tracking server parameters, into a unified AIBoM entity.\\
\textbf{Goals:} Specific SPDX fields assigned to Builders. Goals follow a priority-based fallback framework: if a high-priority Builder fails or returns null, the \textit{PipeManager} automatically invokes the next module in the Team. This tiered execution ensures robustness against API failures or incomplete logs.\\

\textit{\textbf{The LLMClient}}

\label{subsec:llmclient}

To automate the natural-language fields of SPDX 3.0 (such as model limitations), we utilize Large Language Models (LLMs) through a specialized LLMClient. The PipeManager maintains a reference to this client and offers optional generative features to any Builder. 
The LLMClient is established as a formal interface to ensure provider-agnosticism (\eg Ollama, OpenAI). Provides three primary methods: (i)\textit{inferNaturalTextField}: Produces coherent descriptions from logs or code. (ii) \textit{inferFormatted}: Gathers structured data (such as JSON) from unstructured inputs.
(iii) \textit{inferFromImage}: Takes advantage of multimodal capabilities for architecture diagrams or training plots.
Similarly to Builders, the client is created through Java Reflection during the initialization phase. This enables replacing inference providers by supplying a new implementation without altering the core pipeline, thereby ensuring scalability and policy compliance.

\textit{\textbf{Configuration Management}}

AIGen utilizes a two-tier YAML configuration to separate orchestration from the logic specific to each module:\\
\textbf{1. PipeManager Configuration.} Defines the global pipeline state, including:
(i) \textit{Identity \& Compliance:} Metadata for authors and target SPDX versioning;
(ii) \textit{LLM Infrastructure:} Connection details and the specific \textit{LLMClient} implementation to be loaded and
(iii) \textit{Orchestration:} Definition of \textit{Teams}, mapping \textit{Builders} to SPDX components, and assigning \textit{Goals} with execution priorities for fallback.\\
\textbf{2. Builder-Specific Configurations.} Each module (such as MLflow or GitHub) features a tailored schema for platform-specific credentials. A notable aspect is the \textit{focus parameter}, which serves as a semantic signal to the LLM to pinpoint a particular model in multi-model repositories, minimize hallucinations, and improve context accuracy.

This dual-layered method enables organizations to implement concurrent, on-demand pipelines for various model families by simply changing YAML files, ensuring zero-code flexibility to adapt to evolving MLOps frameworks.

\textit{\textbf{Serializer}}

This interface consists of the \textit{init} and \textit{serialize} methods, which distinguish the pipeline from specific output formats by transforming the key-value pairs gathered during the retrieval phase into standardized AIBoM documents.

To keep pace with changing standards (\eg transitioning from SPDX 3.0 to 4.0), the system uses delegation and composition rather than straightforward inheritance. When a new version of a standard is released, fresh abstract builders are created to align with the updated schema.
Specific tool implementations, such as \textit{MLFlowBuilderV4}, act as wrappers for prior versions, allowing the reuse of existing retrieval logic for unchanged fields while implementing only new or modified fields. This approach of ``wrapping'' minimizes the need for extensive refactoring and aids ongoing maintenance in industrial processes as global standards evolve.

\subsection{\textbf{Advanced Data Handling and Robustness}}
The AIBoM Generator employs targeted approaches to manage industrial MLOps artifacts:\\
\textit{\textbf{Notebook Optimization:}} A tailored deserialization layer removes non-essential JSON metadata from Jupyter Notebooks, isolating just the code and markdown to enhance the LLM context window.\\
\textit{\textbf{Sliding Window Extraction:}} To cope with the limited token windows of the used LLMs, the tool uses incremental accumulation. The inputs are split into overlapping segments and processed sequentially to produce a consistent narrative across large files.\\
\textit{\textbf{License Resolution:}} It collects licensing information from various sources, including MLflow registries and GitHub headers. In line with SPDX 3.0, it distinguishes between \textit{Declared} and \textit{Concluded} licenses, employing \texttt{NoAssertion} as the default when the status is unclear to ensure schema adherence.\\
\textit{\textbf{Execution Safety:}} A non-blocking approach prevents pipeline stalls caused by LLM timeouts or reasoning loops. Tasks that surpass limits are automatically re-executed in a ``standard'' (non-reasoning) mode to guarantee the reliability of CI/CD integration.

%% file: implementation.tex
AIGen is developed in Java. The \textit{SPDX 3.0 Serializer} leverages the official Linux Foundation Java library \cite{spdxJavaLib}, specifically utilizing the \textit{AIModel} element. To support industrial requirements for data privacy, the generative backend uses an \textit{OllamaClient} \cite{ollama} to interface with locally hosted LLMs, ensuring that proprietary source code is not exposed to external cloud services.
Regarding the pipeline execution, the \textit{Validation} phase is implemented as a non-blocking audit. Instead of halting the process upon encountering schema inconsistencies, the system logs missing required fields, allowing for subsequent human-in-the-loop refinement without interrupting the automated generation flow.

AIGen follows a configuration-driven usage model. Practitioners provide a pipeline-level YAML file defining the target SPDX profile, the Teams to be instantiated, the Builders assigned to each Team, and the Goals to be populated. Builder-specific configurations provide the connection parameters for external sources, such as MLflow, GitHub, Kaggle, or Hugging Face. At execution time, AIGen initializes the selected Builders, retrieves metadata, serializes it into an SPDX 3.0 JSON-LD AIBoM, and validates the resulting document. New target systems or metadata sources can therefore be handled by replacing configuration files rather than modifying the core implementation. The accompanying YouTube video provides a detailed walkthrough of the configuration structure and command-line usage.

%% file: eval.tex
\begin{figure}[!t]
\centering
\begin{minipage}{\columnwidth}

\begin{lstlisting}[style=jsonstyle, label={lst:ai-package-json}]
  "spdxId": "SpdxRef:model/82201270",
  "type": "ai_AIPackage",
  "suppliedBy": "SpdxRef:creatorAgent/583352542",
  "ai_limitation": "Demonstration model only; not intended for production use\n",
  "ai_informationAboutTraining": "Trained on the Iris dataset using logistic regression\n",
  "ai_useSensitivePersonalInformation": "no",
  "name": "aigen_demo_iris",
  "software_primaryPurpose": "model",
  "software_downloadLocation": "models:/aigen_demo_iris/1",
  "software_packageVersion": "1",
  "ai_hyperparameter": [
    {
      "type": "DictionaryEntry",
      "value": "random_state",
      "key": "42"
    },
    {
      "type": "DictionaryEntry",
      "value": "max_iter",
      "key": "200"
    },
    {
      "type": "DictionaryEntry",
      "value": "dataset",
      "key": "iris"
    },
    {
      "type": "DictionaryEntry",
      "value": "model_family",
      "key": "logistic_regression"
    }
  ]
}
\end{lstlisting}
\end{minipage}
\caption{Excerpt of the AIBoM generated with AIGen.}
\label{fig:excerptAIBOM}
\end{figure}

We performed a preliminary evaluation of AIGen in eight open-source ML projects from GitHub and Kaggle, covering Computer Vision (\eg COVID-xray), Tabular Data (\eg AQI prediction), and Recommendation Systems.


Specifically, we incorporated MLflow into the training scripts of eight open-source projects to track metrics and artifacts. For the inference process, we utilized DeepSeek-r1 8B (Q4\_K\_M quantization) through Ollama, setting the temperature to $0$ and $top\_p=1$ to ensure consistent results. We assessed the generated AIBOM fields using a four-point qualitative scale: Perfect (P) (error-free), Good (G) (minor omissions), Lackluster (L) (substantial missing data), and Bad (B) (factual inaccuracies). We also identified Meta-commentary (M) instances, i.e., instances where the LLM included introspective text. 
Figure~\ref{fig:excerptAIBOM} provides an excerpt of the textual fields generated contained in AIBoM, showing how AIGen synthesizes raw project data into structured SPDX descriptions.

The tool generated 124 AIBoM fields in eight projects, 91\% of the fields were rated Perfect (59\%) or Good (32\%). 


Fields handled by structured Builders, such as MLflow and Kaggle, consistently received Perfect ratings
for deterministic metadata (\ie \textit{Hyperparameters, Metrics}), confirming the effectiveness of the hybrid design when reliable APIs are available.
The most challenging task was the synthesis of descriptive fields from the source code and \texttt{the README} files. In these fields assisted by LLM, we observed that the model often accurately summarized complex training logic when the relevant information was explicit, but occasionally hallucinated or suffered context loss due to the sliding-window strategy. Moreover, domain-specific fields such as \textit{Limitations} often require information beyond the analyzed artifacts, suggesting the need for additional context from domain-specific documentation.

\textbf{AIGen GitHub Repository.} 
The GitHub repository~\cite{replicationpkg} includes an evaluation folder with a list of evaluated projects and field-level quality assessments for AI-generated and metadata from the data set package; the \textit{README} documents the evaluation scope, the rating legend, and the field groups assessed.


%% file: related.tex
This section contextualizes our work within the broader discourse on AI supply chain transparency and evaluates current advancements in AIBoM theoretical frameworks and automation tools.

\textbf{AI in the Software Supply Chain}

Incorporating AI into software development requires improved transparency and traceability. 
Stalnaker \etal \cite{stalnaker2024boms} point out that ensuring content completeness and upkeep is considerably more challenging in machine learning contexts due to complex dependencies and disjointed formats. From an ethical perspective, Widder and Nafus \cite{widder2023dislocated} contend that this fragmentation causes ``dislocated responsibility,'' which could be alleviated through standardized documentation.
Xia \etal \cite{xia2024trust} and Charles \etal \cite{charles2023critical} propose blockchain-based models to safeguard the distribution of SBoM and AIBoM, addressing reliability issues in complex supply chains.

Our research builds on these findings by offering a practical, automated approach for extracting metadata across various MLOps pipelines. While earlier studies emphasize conceptual or infrastructural aspects, our tool aims to fill the operational void by integrating structured data with content generated by large language models, thereby providing concrete assistance for AI governance in industrial environments.

\textbf{Research and Tool Support About AIBoM}

Standard SBoMs fall short in their comprehensiveness for AI systems, often omitting essential datasets, training configurations, and iterative artifacts \cite{xia2023empirical,stalnaker2024boms}. While Xia et al. \cite{xia2024trust} highlight the necessity for dynamic and updatable AIBoMs, practitioners currently do not have access to integrated tools to create them; existing solutions such as DVC \cite{dvcDataVersion} and MLflow \cite{mlflowMLflowMLflow} manage metadata but fail to generate formal, compliant documentation \cite{stalnaker2024boms}. Recently, Rajbahadur et al. \cite{stewartSeip26} established an AIBoM taxonomy as an extension of the SPDX standard (ISO/IEC 5962:2021) to ensure alignment with regulations such as the EU AI Act.

Emerging operational tools focus on different aspects: ALOHA~\cite{d2025aloha} automates the generation of CycloneDX-compliant AIBoMs for pre-trained models from Hugging Face, whereas AIBoMGen~\cite{vandendriessche2026aibomgen} provides verifiable, signed attestations tailored for the training process.

Our research presents a third, complementary approach. In contrast to ALOHA (which focuses on public models) or AIBoMGen (which centers on training integrity), our tool is designed for industrial MLOps settings that are characterized by disjointed data sources. We propose a hybrid pipeline that combines direct extraction from platforms such as MLflow with LLM-based synthesis for unstructured data, fully compliant with the SPDX 3.0 standard, to provide practical documentation support in real-world situations.

%% file: conclusion.tex
We introduced AIGen, a modular tool that automates the creation of SPDX‑compliant AIBoMs by combining structured MLOps extraction with LLM‑based synthesis. The architecture supports extensibility, evolving standards, and heterogeneous pipelines. Preliminary results on eight projects show high accuracy for structured fields and highlight remaining challenges in natural‑language synthesis. 
Future work will extend the current open-source evaluation with an industrial validation conducted in collaboration with our partner company. In particular, we will apply AIGen to real MLOps pipelines to assess its configurability, robustness against incomplete metadata, and usefulness in supporting human-in-the-loop auditing of compliance-relevant AI documentation.